%% file: iclr2027_conference.tex
\documentclass{article} % For LaTeX2e
\usepackage{iclr2027_conference,times}

\input{math_commands.tex}

\usepackage{hyperref}
\usepackage{url}
\usepackage{graphicx}
\usepackage{amsmath,amssymb,booktabs}
\usepackage{pifont}
\usepackage{placeins}
\usepackage{capt-of}

\title{Exploring Temporal Context-Aware Physical Grounding in Latent World Models}

\author{
\textbf{Yilei Man}$^{1}$,
\textbf{Yang Chen}$^{1}$,
\textbf{Jingcai Guo}$^{1,2}$\thanks{Jingcai Guo is the corresponding author.}\\
$^{1}$Department of Computing, The Hong Kong Polytechnic University\\
$^{2}$Department of Land Surveying and Geospatial Science, The Hong Kong Polytechnic University\\
\texttt{jc-jingcai.guo@polyu.edu.hk}
}

\iclrfinalcopy % Uncomment only for the camera-ready version.
\begin{document}
\raggedbottom % Keep vertical spacing natural.

\maketitle
\suppressfloats[t] % Keep the introduction teaser below the title on page 1.

\begin{abstract}

Learning latent dynamics is fundamental to building action-conditioned world models for planning. Among existing families, Joint Embedding Predictive Architectures (JEPAs) offer a natural framework by modeling transitions directly in latent space rather than at the pixel level. However, two issues remain in JEPA-based world models: (i) physical grounding lacks explicit aggregation of temporal context and (ii) action planning always overly relies on generic sampling distributions. To this end, we propose $\mathcal{G}$-JEPA, which incorporates temporal context into physical grounding for latent world models and couples the learned models with prior-guided action planning. Specifically, during learning, we incorporate a shared causal temporal module that aggregates temporal context within observation and prediction sequences under physical state supervision, grounding both in task-relevant physical states. When planning, we further introduce action inference through stepwise conditioning of a generative action distribution on the current and goal latent representations together with their evolving discrepancy, yielding anchored proposals that guide model-based online refinement. Together, physical state supervision guides temporal representation learning, while the action prior guides search through the learned world model. Experiments demonstrate improved planning performance on manipulation tasks, including tasks with increasing temporal separation between initial and goal observations. Code is publicly available at \url{https://github.com/myl11045/G-JEPA}.
\end{abstract}

\section{Introduction}

World models learn action-conditioned environment dynamics, enabling agents to anticipate and evaluate the consequences of candidate actions. For learning from visual observations, an important design choice is whether to generate future observations or predict their representations. Generative approaches synthesize future observations directly in pixel space \citep{lee2025edeline} or through latent representations learned for visual reconstruction \citep{huang2026vid2world}, requiring the preservation of appearance information beyond what may be necessary for control. Representation-predictive approaches instead learn action-conditioned dynamics by predicting the embeddings of future observations \citep{zhou2025dinowm,assran2025vjepa2}. Removing the reconstruction requirement offers the potential to learn representations that emphasize task-relevant dynamics over incidental visual detail.

\begin{figure}[!htb]
    \centering
    \includegraphics[width=\linewidth]{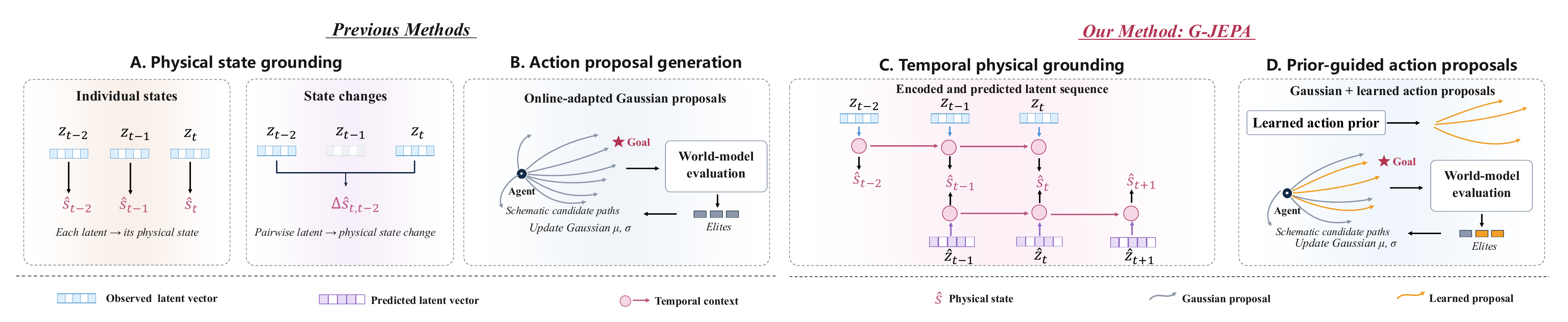}
    \caption{\textbf{Temporal physical grounding and prior-guided planning.}
    \textbf{(A)} PhyLatent \citep{zeng2025phylatent} grounds individual latents in physical states, while PSG-JEPA \citep{yan2026psgjepa} additionally supervises state changes between latent pairs.
    \textbf{(B)} Gaussian CEM generates action candidates from a proposal distribution adapted through online elite updates.
    \textbf{(C)} $\mathcal{G}$-JEPA uses a shared causal grounding module to aggregate temporal context within encoded and predicted latent sequences under physical state supervision.
    \textbf{(D)} Learned action proposals supplement Gaussian
    candidates, with joint world-model evaluation and shared elite selection guiding subsequent Gaussian updates.
    Curves illustrate schematic candidate motion paths.}
    \label{fig:temporal_grounding_teaser}
\end{figure}

Joint-Embedding Predictive Architectures (JEPAs) provide a natural framework for this approach by predicting target embeddings from context representations without reconstructing the observations \citep{assran2023ijepa,bardes2024vjepa}. World models such as DINO-WM and V-JEPA 2-AC train action-conditioned predictors on frozen DINOv2 and video-pretrained JEPA features, respectively \citep{oquab2023dinov2,zhou2025dinowm,assran2025vjepa2}. Although visual pretraining provides strong perceptual priors, freezing the encoders prevents interaction data from shaping representations around the task-specific physical structure needed for control and policy learning. LeWorldModel (LeWM) trains its visual encoder and action-conditioned predictor end to end from raw pixels using only future-embedding prediction and a distributional regularizer \citep{maes2026lewm}. Its predictive objective aligns predicted embeddings with those of future observations, but does not explicitly identify or prioritize control-relevant physical information. 

Recent approaches introduce physical supervision to make this
information explicit. PhyLatent \citep{zeng2025phylatent} grounds individual latents in physical states, while PSG-JEPA \citep{yan2026psgjepa} also supervises state changes between latent pairs (Figure~\ref{fig:temporal_grounding_teaser} A). Physical state estimation can also draw on information distributed across time, but these grounding heads do not explicitly accumulate this temporal context. For action selection, many JEPA-based world models \citep{zhou2025dinowm,assran2025vjepa2,maes2026lewm} adopt model predictive control with the cross-entropy method (CEM). Conventional CEM refines a generic proposal distribution through online elite updates, without directly incorporating behavioral priors learned from offline trajectories (Figure~\ref{fig:temporal_grounding_teaser} B). Under a limited candidate budget, this can leave task-relevant action sequences insufficiently represented in the candidate set.

% To this end, we propose \textbf{$\mathcal{R}$-JEPA}, which combines recursive physical grounding during world-model learning with prior-guided recursive planning. Its \textbf{Quasi-Linear Physical Grounding (QLPG)} module uses a shared causal S4D network to aggregate temporal context into fixed-size recurrent memory with linear complexity in sequence length (Figure~\ref{fig:temporal_grounding_teaser}(c)). The observation branch learns to estimate task-relevant physical states from this memory. The prediction branch inherits the memory and processes action-conditioned predicted latents through the same state update and physical state estimator, with supervision from the corresponding future physical states. Nuisance-invariant representation learning further reduces sensitivity to appearance variations that preserve physical states.

To this end, we propose \textbf{$\mathcal{G}$-JEPA}, which incorporates temporal context into physical grounding for latent world models and combines the learned models with prior-guided action planning. The shared temporal physical grounding module is built on S4D \citep{gu2022s4d}, a diagonal structured state-space model. Its observation and prediction branches separately aggregate temporal context from encoded observations and action-conditioned predicted latents through causal updates to estimate the corresponding physical states (Figure~\ref{fig:temporal_grounding_teaser} C). Physical state supervision on both branches complements latent matching, guiding the encoder and predictor to retain task-relevant physical information. 
% This temporal aggregation uses fixed-size recurrent memory with computational complexity linear in sequence length. 
% S4D supports this temporal aggregation with fixed-size recurrent memory and linear complexity in sequence length. 
Nuisance-invariant representation learning further reduces sensitivity to appearance variations that preserve physical states. We then freeze the world model and learn a conditional generative action prior from offline trajectories to propose action sequences based on the current and goal latent representations. 
% We then freeze the world model and train a latent-conditioned diffusion policy to generate action proposals. 
These proposals are mixed with Gaussian CEM candidates and evaluated using the same latent objective (Figure~\ref{fig:temporal_grounding_teaser} D). A shared elite set updates the Gaussian search distribution, connecting the learned action prior with online refinement. After action execution, the updated observation conditions the next round of proposal generation and planning.

Our contributions are:
\begin{itemize}
    \item We introduce temporal physical grounding for JEPA world models. A shared causal S4D module aggregates context within observation and prediction sequences, using physical state supervision to jointly guide representation learning and latent prediction.

    % \item We integrate learned latent-conditioned action proposals with generic CEM. Unified world-model scoring and shared elite updates connect offline behavioral priors with model-based online refinement.
    \item We develop prior-guided action planning that integrates a learned latent-conditioned action prior with Gaussian CEM. Unified world-model scoring and shared elite updates connect offline behavioral priors with model-based online refinement.

    \item We evaluate $\mathcal{G}$-JEPA on PushT and OGBench Cube through planning comparisons and ablations of physical grounding and action selection, with additional PushT evaluations at increasing goal distances. The results support the effectiveness of combining temporal physical grounding with prior-guided action proposals for JEPA planning.
\end{itemize}

% Planning also requires action proposals that adapt to the current state and goal. We introduce a diffusion policy conditioned on their latent representations to generate goal-directed action sequences. These proposals are combined with CEM samples and evaluated through world-model predictions to select actions for execution. After execution, the updated observation conditions the next round of proposal generation and planning. This closed-loop process recursively adapts action selection as the environment evolves toward the goal.

% Our contributions are:
% \begin{itemize}
%     \item We introduce recursive physical grounding through a shared S4D module. By inheriting memory fro observation histories, the prediction branch extends physical state grounding to action-conditioned future latents.

%     \item We develop prior-guided recursive planning that combines diffusion-generated action proposals with CEM search. Updated observations guide successive planning rounds, connecting learned action generation with closed-loop control.

%     \item We evaluate $\mathcal{R}$-JEPA on PushT and Cube through planning comparisons and ablations of physical grounding and action selection, with additional PushT evaluations at increasing goal distances. The results support the effectiveness of combining recursive physical grounding with learned action proposals for JEPA planning.
% \end{itemize}

\section{Related Work and Background}
\label{sec:related_work}

\subsection{Joint Embedding Predictive Architectures}
\label{sec:related_world_models}

JEPA has been applied to image representation learning with I-JEPA \citep{assran2023ijepa}, video understanding with V-JEPA \citep{bardes2024vjepa}, vision-language tasks with
VL-JEPA \citep{chen2025vljepa}, and robotic control with VLA-JEPA \citep{sun2026vlajepa}. Building on V-JEPA 2 \citep{assran2025vjepa2}, pretrained on over one million hours of internet video, V-JEPA 2.1 \citep{murlabadia2026vjepa21} improves fine-grained visual understanding for tasks such as depth estimation, object tracking, and robotic planning. LeWM \citep{maes2026lewm} introduces a lightweight JEPA world model trained end to end from pixels for action-conditioned prediction and planning. Subsequent work incorporates physical supervision, temporal progress, and online adaptation to strengthen these predictive models. PhyLatent \citep{zeng2025phylatent} combines physical state grounding of encoded and predicted latents with appearance invariance and counterfactual dynamics constraints. PSG-JEPA \citep{yan2026psgjepa} grounds individual latents in robot proprioceptive states and latent pairs in multi-horizon joint-angle changes. Beyond physical state supervision, TD-JEPA \citep{bai2026temporaldistancejepa} learns directed temporal progress from offline trajectories to shape representations and planning costs, while AdaJEPA \citep{wang2026adajepa} adapts the world model using newly observed transitions during planning. Our work introduces temporal aggregation into physical grounding
and integrates learned action proposals with model-based online
refinement.

% Our work extends physical grounding through recurrent memory inherited from observation histories to predicted futures, and integrates learned action proposals with CEM in a recursive planning loop.

\subsection{Structured State-Space Models}
\label{sec:ssm_background}

Structured state-space models provide a framework for modeling temporal dependencies with compact recurrent memory. HiPPO \citep{gu2020hippo} formulates online history compression through projections onto polynomial bases. S4 \citep{gu2022s4} develops structured state-space layers for long-range sequence modeling, while S4D \citep{gu2022s4d} simplifies their parameterization through diagonal state transitions. More recent architectures, including Mamba-2 \citep{dao2024mamba2}, Gated DeltaNet \citep{yang2025gateddeltanet}, and Gated DeltaNet-2 \citep{hatamizadeh2026gateddeltanet2},
further explore efficient sequence modeling with recurrent memory. We use a shared S4D module to aggregate temporal context separately within observation and prediction sequences for physical state estimation.
% We use S4D to maintain recurrent memory across observation histories and predicted latent sequences under physical state supervision.

\subsection{Diffusion Policies and Model-Based Planning}
\label{sec:related_diffusion_planning}

Diffusion models have been used for both action generation and trajectory planning. Diffusion Policy \citep{chi2024diffusionpolicy} learns observation-conditioned action sequences for robotic manipulation, while Diffuser  \citep{janner2022diffuser} generates trajectories guided by goals or rewards. Further work combines diffusion-generated candidates with online planning. D-MPC \citep{zhou2025diffusionmpc} learns separate diffusion models for action proposals and future dynamics, using predicted outcomes to select actions within an MPC loop. Diffusion-ES \citep{yang2024diffusiones} combines trajectory generation with evolutionary search for autonomous driving. More closely related to hybrid planning, MODIP \citep{elasri2026modip} mixes diffusion proposals with Gaussian candidates for latent model predictive control and uses the resulting trajectories to improve the policy. Our approach combines proposals from a learned latent-conditioned action prior with Gaussian CEM candidates, evaluated by a physically grounded JEPA world model using a latent goal-matching objective. The action prior remains fixed, while a shared elite set updates the Gaussian sampling distribution for online refinement.
% Our approach combines latent-conditioned diffusion proposals with Gaussian CEM candidates, evaluated by a physically grounded JEPA world model using a latent goal-matching objective. The diffusion policy remains fixed, while a shared elite set updates the generic sampling distribution for online refinement.
% Our approach combines latent-conditioned diffusion proposals with Gaussian CEM candidates, evaluated by a physically grounded JEPA world model using a latent goal-matching objective. The policy remains fixed and is reconditioned on updated observations at each planning round.

\section{Methodology}
\label{sec:method}
\label{sec:method_overview}

\begin{figure}[t]
    \centering
    \includegraphics[width=\linewidth]{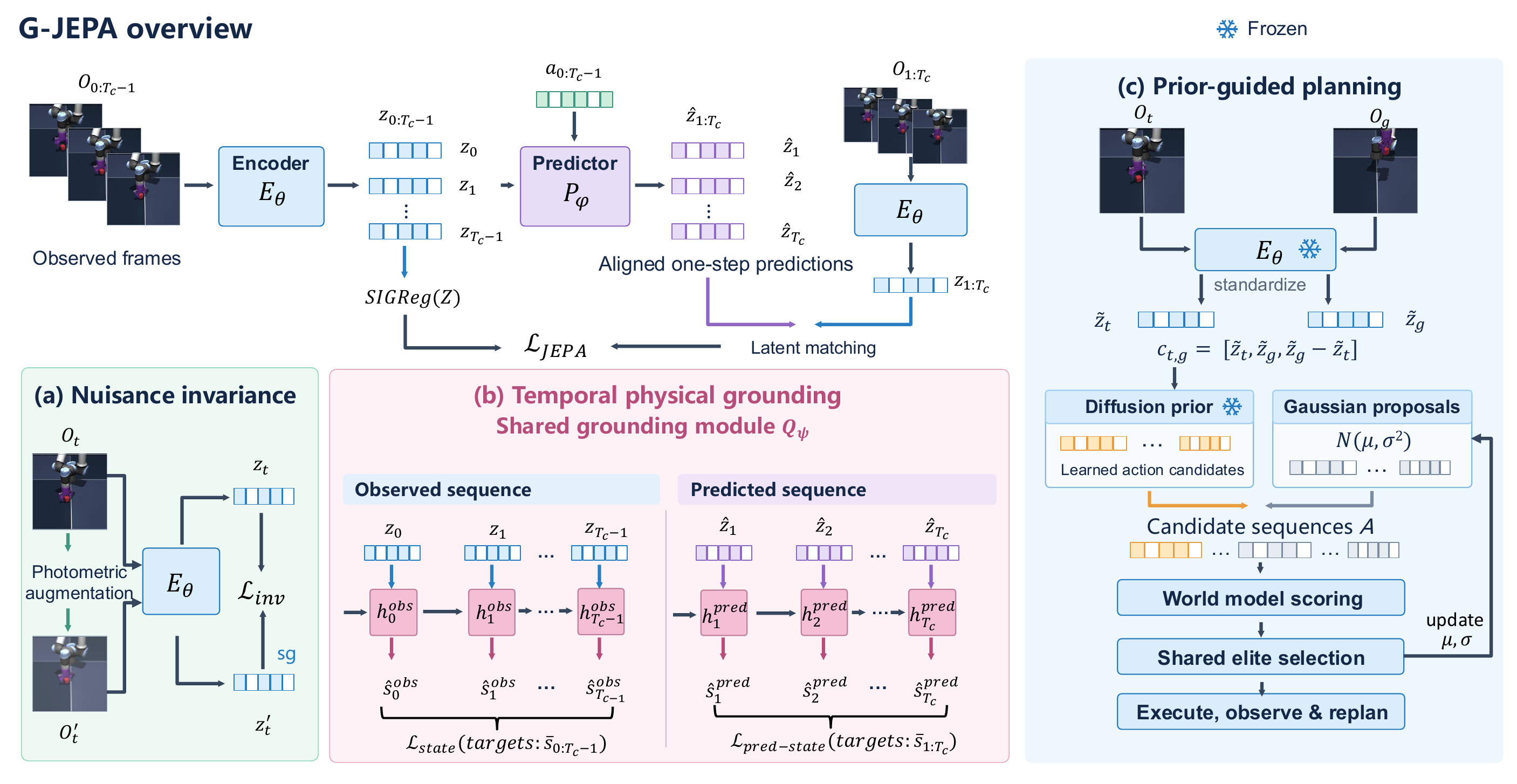}%
    \caption{\textbf{Overview of $\mathcal{G}$-JEPA.} World-model learning combines latent matching and SIGReg with
(a) nuisance invariance under photometric augmentation and
(b) temporal physical grounding.
A shared S4D module aggregates temporal context separately
within observed latents and aligned one-step predictions,
supervising physical state estimates from both sequences.
(c) During planning, the learned action prior is instantiated
as a latent-conditioned diffusion policy that generates
action proposals alongside Gaussian candidates.
The frozen world model scores both sources using latent
goal-matching costs, and a shared elite set updates the
Gaussian search distribution.
The policy remains fixed, while action execution provides
new observations for subsequent planning rounds.}
    \label{fig:gjepa_overview}
\end{figure}

In this section, we introduce $\mathcal{G}$-JEPA (Figure~\ref{fig:gjepa_overview}). We first present the necessary preliminaries. Then, we introduce a shared causal module for temporal physical grounding of observation and prediction sequences. Finally, we present prior-guided planning that combines latent-conditioned diffusion proposals with the generic CEM search.

% Then, we learn a physically grounded recursive dynamics memory bridging observed and imagined latents. Finally, we present recursive action inference with stepwise-conditioned proposals for latent planning.

% $\mathcal{R}$-JEPA integrates recursive physical grounding
% and prior-guided planning into end-to-end JEPA world modeling.
% QLPG extends physical grounding from observation histories
% to predicted latents through shared S4D recurrence and
% memory inheritance. Nuisance-invariant representation learning further reduces sensitivity to appearance variations that preserve physical states. We then freeze the world model and learn a diffusion action prior, which guides CEM search within a closed-loop planning process.

\subsection{LeWorldModel Preliminary}
\label{sec:problem}

% \textbf{LeWM Formulation.}
LeWorldModel \citep{maes2026lewm} consists of an encoder $E_\theta$ and an action-conditioned predictor $P_\varphi$. A trajectory segment contains $T+1$ image observations $\mathbf{o}_{0:T}$ and $T$ actions $\mathbf{a}_{0:T-1}$, where $\mathbf{a}_t$ is executed between $\mathbf{o}_t$ and $\mathbf{o}_{t+1}$. The encoder maps each observation $\mathbf{o}_t$ to a latent representation $\mathbf{z}_t=E_\theta(\mathbf{o}_t)$. Conditioned on the current latent and action, the predictor estimates the latent representation of the subsequent observation as $\hat{\mathbf{z}}_{t+1}=P_\varphi(\mathbf{z}_t,\mathbf{a}_t)$. For a history of $N$ observations, the resulting latent representations are stacked into $\mathbf{Z}\in\mathbb{R}^{N\times d}$, where $d$ denotes the latent dimensionality. An anti-collapse regularizer $\mathrm{SIGReg}(\mathbf{Z})$ is applied to encourage feature diversity. The overall training objective is then defined as follows:
\begin{equation}
    \mathcal{L}_{\mathrm{JEPA}}
    =
    \frac{1}{N}
    \sum_{t=0}^{N-1}
    \left\|\hat{\mathbf{z}}_{t+1}-\mathbf{z}_{t+1}\right\|_2^2 + \lambda_{\mathrm{reg}}\mathrm{SIGReg}(\mathbf{Z}).
    \label{eq:jepa_training}
\end{equation}

\subsection{Learning in Observed and Imagined Latent Space}
\label{sec:temporal_grounding}
\label{sec:grounding_objectives}

% To maintain latent state consistency across the observed and imagined space, in this section, we incorporate a recursive memory mechanism to learn its physical dynamics. 
\textbf{Observed Physical State Grounding.}
To incorporate temporal context beyond individual latents
or isolated latent pairs, the observation branch causally
aggregates encoded observations to estimate task-relevant
physical states. Let $\mathbf{s}_t$ denote the ground-truth physical state and $\bar{\mathbf{s}}_t$ its normalized form, with dimension $d_s$. The shared grounding module $Q_\psi$ (Figure~\ref{fig:gjepa_overview}(b)) consists of two stacked S4D-based layers \citep{gu2022s4d}, followed by a nonlinear physical state estimator. 
% We denote the recurrent memories of both layers by $\mathbf{h}_t=(\mathbf{h}_t^{(1)},\mathbf{h}_t^{(2)})$. 
The observation branch updates its memory and estimates the physical state as
\begin{equation}
    (\hat{\mathbf{s}}_t^{\mathrm{obs}},\mathbf{h}_t^{\mathrm{obs}})
    =
    Q_\psi(\mathbf{z}_t,\mathbf{h}_{t-1}^{\mathrm{obs}}),
    \qquad \mathbf{h}_{-1}^{\mathrm{obs}}=\mathbf{0}.
    \label{eq:observed_grounding}
\end{equation}
Here $\hat{\mathbf{s}}_t^{\mathrm{obs}}$ is the physical state estimate in the same normalized space as $\bar{\mathbf{s}}_t$, and $\mathbf{h}_t^{\mathrm{obs}}$ contains the updated recurrent memories. 
% The superscript $\mathrm{obs}$ identifies the observation branch.
Within each S4D layer, the memory is updating as follows:
% $\mathbf{h}_t^{(\ell)}\in\mathbb{R}^{N}$ 
% \begin{equation}
%     \mathbf{h}_t^{(\ell)}
%     =
%     \boldsymbol{\alpha}\odot \mathbf{h}_{t-1}^{(\ell)}
%     + B_\ell \mathbf{v}_t^{(\ell)},
%     \label{eq:temporal_memory}
% \end{equation}
\begin{equation}
    \mathbf{h}_t = \boldsymbol{\alpha}\odot \mathbf{h}_{t-1} + B\mathbf{v}_t,
    \label{eq:temporal_memory}
\end{equation}
where 
% $\ell\in\{1,2\}$ indexes the layer, $N$ is its memory dimension, and 
$\odot$ denotes elementwise multiplication. The normalized input $\mathbf{v}_t$ enters the memory through a learned projection $B$. 
The 
% vector 
$\boldsymbol{\alpha}=(\alpha_0,\ldots,\alpha_{N_{\alpha}-1})^\top$ contains fixed decay coefficients:
\begin{equation}
    \alpha_i
    =
    \exp\!\left[-\Delta\left(i+\tfrac{1}{2}\right)\right],
    \qquad i=0,\ldots,N_{\alpha}-1,
    \label{eq:temporal_memory_coefficient}
\end{equation}
where $N_\alpha$ is the memory dimension and $\Delta>0$
controls the decay rate. Larger coefficients preserve historical contributions longer, while smaller coefficients emphasize recent inputs. A learned output projection transforms each layer's memory into features for the next layer or the physical state estimator. For fixed model dimensions, this recurrence maintains a fixed-size hidden state and requires computation linear in sequence length.

We ground the observation representations by matching the estimated physical states to their normalized targets:
\begin{equation}
    \mathcal{L}_{\mathrm{state}}
    =
    \frac{1}{T_{C}d_s}
    \sum_{t=0}^{T_{C}-1}
    \left\|\hat{\mathbf{s}}_t^{\mathrm{obs}}-\bar{\mathbf{s}}_t\right\|_2^2,
    \label{eq:observed_state_loss}
\end{equation}
where $T_{C}$ is the observation history length.

\textbf{Predicted Physical State Grounding.}
% Latent matching measures discrepancies between embeddings without directly evaluating errors in physical state estimation. We therefore apply the shared grounding module to the predicted latent sequence and supervise its state estimates, extending physical supervision from encoded observations to latent predictions. The JEPA predictor $P_\varphi$ generates aligned one-step predictions $\hat{\mathbf{z}}_{1:T_C}$ from the observation context and recorded actions \citep{maes2026lewm}. The prediction branch initializes its recurrent memory to $\mathbf{h}_0^{\mathrm{pred}}=\mathbf{0}$. Using the shared grounding module $Q_\psi$, it aggregates temporal context within the predicted latent sequence and estimates the corresponding physical states:
Following LeWM \citep{maes2026lewm}, the predictor $P_\varphi$
generates aligned one-step predictions $\hat{\mathbf{z}}_{1:T_C}$ from the observation context and recorded actions. To supervise these predictions in physical state space, we process this sequence with the shared grounding module $Q_\psi$. Starting from $\mathbf{h}_0^{\mathrm{pred}}=\mathbf{0}$, the module aggregates temporal context and estimates the corresponding physical states:

\begin{equation}
(\hat{\mathbf{s}}_k^{\mathrm{pred}},
 \mathbf{h}_k^{\mathrm{pred}})
=
Q_\psi(
\hat{\mathbf{z}}_k,
\mathbf{h}_{k-1}^{\mathrm{pred}}),
\qquad k=1,\ldots,T_C.
\label{eq:predicted_grounding}
\end{equation}
The superscript $\mathrm{pred}$ identifies the prediction branch.
We supervise the predicted physical states using their
corresponding normalized targets:
\begin{equation}
\mathcal{L}_{\mathrm{pred\text{-}state}}
=
\frac{1}{T_Cd_s}
\sum_{k=1}^{T_C}
\left\|
\hat{\mathbf{s}}_k^{\mathrm{pred}}
-
\bar{\mathbf{s}}_k
\right\|_2^2.
\label{eq:physical_grounding}
\end{equation}

\textbf{Nuisance-Invariant Physical State.}
To reduce the influence of appearance-related nuisance factors on physical-state recovery, we encourage the encoder to learn representations that are invariant to photometric perturbations (Figure~\ref{fig:gjepa_overview}(a)). Specifically, for each observation $\mathbf{o}_t$, we construct an augmented view $\mathbf{o}'_t$ by randomly adjusting its brightness, color, and saturation and adding Gaussian noise in pixel space. We encode the original and augmented observations as $\mathbf{z}_t=E_\theta(\mathbf{o}_t)$ and $\mathbf{z}'_t=E_\theta(\mathbf{o}'_t)$, respectively, and define the following invariance objective:
\begin{equation}
    \mathcal{L}_{\mathrm{inv}}=\frac{1}{T_c+1}\sum_{t=0}^{T_c}
    \left[
        1-\operatorname{cos}
        \bigl(\mathbf{z}_t,\operatorname{sg}(\mathbf{z}'_t)\bigr)
    \right],
    \label{eq:observation_invariance}
\end{equation}
where $\operatorname{cos}(\cdot,\cdot)$ denotes cosine similarity and
$\operatorname{sg}(\cdot)$ denotes the stop-gradient operator.

\textbf{Training Objective.}
% We use two-stage training. 
% First, 
We jointly optimize the visual encoder, action-conditioned predictor, and grounding module as follows:
\begin{equation}
\begin{aligned}
\mathcal{L} ={}&
\mathcal{L}_{\mathrm{JEPA}}
+ \lambda_s\mathcal{L}_{\mathrm{state}} + \lambda_p\mathcal{L}_{\mathrm{pred\text{-}state}}
+ \lambda_i\mathcal{L}_{\mathrm{inv}}.
\end{aligned}
\label{eq:total_objective}
\end{equation}
% Here $\mathcal{L}_{\mathrm{SIGReg}}$ denotes the representation regularizer used in LeWM \citep{maes2026lewm}. 
We set $\lambda_{\mathrm{reg}}=\lambda_s=0.09$ and $\lambda_p=\lambda_i=0.05$.
% Second, we freeze the world model and train the diffusion policy by minimizing $\mathcal{L}_{\mathrm{act}}$.

\subsection{Prior-Guided Latent Planning}
\label{sec:action_diffusion}

\textbf{Action Prior Acquisition.} 
A latent-conditioned generative action prior is learned from offline trajectories using latent representations produced by the frozen encoder. Each training sample pairs a current image
$\mathbf{o}_t$ with a goal image $\mathbf{o}_g$ located $K$
action chunks later in the same offline trajectory.
The intervening action sequence
$\mathbf{A}=(\mathbf{a}_t,\ldots,\mathbf{a}_{t+K-1})$
provides supervision.
The frozen encoder maps both images to latent embeddings.
These embeddings are standardized using the mean and standard
deviation of the policy-training embeddings, yielding
$\tilde{\mathbf{z}}_t$ and $\tilde{\mathbf{z}}_g$.
The conditioning input is
\begin{equation}
\mathbf{c}_{t,g}
=
[\tilde{\mathbf{z}}_t,\tilde{\mathbf{z}}_g,
 \tilde{\mathbf{z}}_g-\tilde{\mathbf{z}}_t],
\label{eq:goal_condition}
\end{equation}
where the brackets denote feature concatenation.
% The prior models the conditional action distribution $\pi_{\mathrm{prior}}(\mathbf{A}\mid\mathbf{c}_{t,g})$, from which planning proposals are sampled.
The learned action prior defines a distribution $\pi_{\mathrm{prior}}(\mathbf{A}\mid\mathbf{c}_{t,g})$ over candidate action sequences $\mathbf{A}$. Sampling from this distribution provides action proposals for planning.
% The prior models the conditional action distribution $\pi_{\mathrm{prior}}(\mathbf{A}\mid\mathbf{c}_{t,g})$, from which candidate action sequences are sampled for planning.
In this work, the prior is instantiated as a diffusion policy
following Diffusion Policy \citep{chi2024diffusionpolicy},
conditioned on $\mathbf{c}_{t,g}$.
The policy is trained with a noise-prediction objective
$\mathcal{L}_{\mathrm{act}}$, defined as the mean squared error
between the predicted noise and the Gaussian noise added
to the recorded action sequences.

% First, we train a latent-conditioned diffusion policy to provide action priors for planning \citep{chi2024diffusionpolicy}.
% It is trained on frozen representations learned with QLPG. Each training sample pairs a current image $\mathbf{o}_t$ with a goal image $\mathbf{o}_g$ located $K$ action chunks later in the same offline trajectory. The intervening action sequence $\mathbf A=(\mathbf{a}_t,\ldots,\mathbf{a}_{t+K-1})$ provides supervision.
% The frozen encoder maps both images to latent embeddings. We standardize them using the mean and standard deviation of the policy-training embeddings, yielding $\tilde{\mathbf{z}}_t$ and $\tilde{\mathbf{z}}_g$. The policy condition is
% \begin{equation}
%     \mathbf{c}_{t,g}
%     =
%     [\tilde{\mathbf{z}}_t,\tilde{\mathbf{z}}_g,\tilde{\mathbf{z}}_g-\tilde{\mathbf{z}}_t],
%     \label{eq:goal_condition}
% \end{equation}
% where the brackets denote feature concatenation. The policy is trained with a noise-prediction objective
% $\mathcal{L}_{\mathrm{act}}$, defined as the mean squared
% error between the predicted noise and the Gaussian noise
% added to the recorded action sequences.
% % The policy learns the conditional action distribution through a noise-prediction objective $\mathcal L_{\mathrm{act}}$.
% % At inference, the policy samples complete action sequences conditioned on $\mathbf{c}_{t,g}$. These serve as \textbf{amortized global proposals}: offline learning supplies a structured action prior, while candidate generation is not restricted by the current CEM mean or variance.

\textbf{Model-Based Online Refinement.}
Planning uses mixed-proposal CEM to combine the learned
action prior with online refinement guided by the physically
grounded world model (Figure~\ref{fig:gjepa_overview}(c)). At each planning round, the latest observation and the fixed goal define $\mathbf{c}_{t,g}$ to condition the action prior for proposal generation. At each CEM iteration, proposals sampled from $\pi_{\mathrm{prior}}(\cdot\mid\mathbf{c}_{t,g})$ occupy $\rho N_{\mathrm{cand}}$ slots in a fixed population of $N_{\mathrm{cand}}$ candidates. The remaining slots contain samples from the evolving CEM Gaussian and its retained mean candidate.

Let $K$ denote the number of model steps in each planning
rollout. For each candidate sequence $\mathbf{A}$, the frozen predictor recursively generates future embeddings $\hat{\mathbf{z}}_{t+k|t}$ under its action chunks. Given the goal embedding $\mathbf{z}_g=E_\theta(\mathbf{o}_g)$, we rank candidates from both proposal sources using
\begin{equation}
J(\mathbf{A})
=
\sum_{k=1}^{K} w_k
\operatorname{MSE}\!\left(
\hat{\mathbf{z}}_{t+k|t},\mathbf{z}_g
\right),
\label{eq:planning_objective}
\end{equation}
where $w_k\geq 0$ is the weight assigned to prediction step $k$.

The lowest-cost candidates form a shared elite set,
whose statistics update the Gaussian mean and variance. Prior-generated candidates entering this set guide subsequent
Gaussian search, coupling learned action proposals with
model-based online refinement. Fresh samples from the learned action prior are introduced at every CEM iteration, while its parameters remain fixed.

After optimization, we execute the prescribed action segment
and obtain a new observation. This observation updates the conditioning input to the action prior and initiates the next planning round.

\section{Experiments}
\label{sec:experiments}

We evaluate temporal physical grounding and prior-guided action planning through cross-method comparisons and controlled ablations. 
% PushT provides the main comparisons, OGBench Cube evaluates robot manipulation, and TwoRooms provides a navigation reference.

\subsection{Experimental Setup}
\label{sec:experimental_setup}

\paragraph{Tasks and data.}
We consider PushT manipulation, OGBench Cube pick-and-place \citep{park2025ogbench}, and TwoRooms navigation using LeWM's visual datasets \citep{maes2026lewm}. World-model training uses a 90/10 training--validation split at the window level for PushT and at the episode level for OGBench Cube and TwoRooms. Grounding targets are seven-dimensional agent--block states for PushT, 18-dimensional robot--object states for OGBench Cube, and two-dimensional agent positions for TwoRooms.

For PushT planning, we use reference tasks and tasks
with increasing goal distances. \textbf{PushT-Ada} contains 256 fixed tasks from AdaJEPA's planning pool \citep{wang2026adajepa}. \textbf{PushT-Distance} follows the goal-distance evaluation in HWM \citep{zhang2026hwm}, which extends DINO-WM's goal-conditioned PushT setting \citep{zhou2025dinowm}. We construct this set from validation windows spanning 256 distinct episodes. Each episode provides a shared goal and starting points 25, 50, and 75 recorded environment steps before that goal, yielding 256 tasks per distance. \textbf{PushT-Challenging} is a more challenging subset of PushT-Distance. It contains 126, 209, and 240 tasks at the respective distances. We also evaluate planning in robot manipulation and navigation. \textbf{OGBench Cube} uses TD-JEPA's fixed 50-task reference set \citep{bai2026temporaldistancejepa}, and \textbf{TwoRooms} uses 100 tasks drawn from validation episodes. Data preparation and task construction are detailed in Appendix~\ref{app:data_tasks}.

\paragraph{Models and training.}
$\mathcal{G}$-JEPA builds on LeWM with a ViT-tiny encoder and 192-dimensional latent representations. The temporal physical grounding module uses two S4D-based layers with memory dimension $N_\alpha=64$ and fixed decay scale $\Delta=0.1$. The observation context and latent-prediction setup follow LeWM's training protocol.
World models are trained on NVIDIA H200 GPUs for PushT and TwoRooms, and NVIDIA H100 GPUs for OGBench Cube. Each main configuration uses one training seed; the temporal-module comparison uses three training replicates. LeWM, TD-JEPA, and PSG-JEPA are retrained and evaluated for comparison under common dataset-specific planning protocols.
World model optimization and baseline adaptations are detailed in Appendix~\ref{app:world_training}, and diffusion-policy training in Appendix~\ref{app:diffusion_training}.

\paragraph{Planning and metrics.}
Planning uses latent goal-matching costs computed from the frozen world model. The temporal physical grounding module is used only during training. The full planner combines action diffusion (AD) proposals with Gaussian CEM search for model-based online refinement. We use $\rho=0.2$ for $\mathcal{G}$-JEPA on PushT-Ada, OGBench Cube, PushT-Distance, and PushT-Challenging, following the proposal-ratio sweep in Figure~\ref{fig:diffusion_ratio}. Without AD, $\rho=0$ and planning uses Gaussian CEM alone. For PushT-Ada, OGBench Cube, and TwoRooms, the planning horizon is 25 environment steps ($K=5$). Each model step spans five environment steps and is conditioned on the corresponding five-action chunk. PushT-Ada follows AdaJEPA's temporal weighting rule \citep{wang2026adajepa}, while OGBench Cube and TwoRooms use terminal latent matching throughout. All methods follow the dataset-specific comparison protocols in Appendix~\ref{app:protocols}. Table~\ref{tab:base_planning_protocols} in Appendix~\ref{app:protocols} lists execution budgets, success thresholds, and diffusion sampling settings.
Success is measured by reaching the goal within the execution budget. We report mean success rates (\%) over three planning seeds for PushT and TwoRooms and ten for OGBench Cube. In tables, subscripts indicate $\pm$ one sample standard deviation across planning seeds, or across training replicates for the temporal-module comparison.

\subsection{Performance Comparison}
\label{sec:main_planning}

\paragraph{Cross-method comparison.}
Table~\ref{tab:cross_dataset} compares $\mathcal{G}$-JEPA with
LeWM, TD-JEPA, and PSG-JEPA across datasets under matched
planning budgets.
On PushT-Ada, $\mathcal{G}$-JEPA without AD reaches $79.43\%$
success, exceeding LeWM's $56.90\%$ and TD-JEPA's $56.64\%$.
Adding AD further raises success to $87.11\%$, showing that
learned action proposals complement the grounded world model
on this task.
On OGBench Cube, $\mathcal{G}$-JEPA without AD achieves
$78.60\%$, comparable to TD-JEPA's $78.00\%$ and above LeWM's
$72.80\%$. Adding AD raises success to $100\%$, demonstrating
a substantial gain from incorporating learned proposals
into planning.
On TwoRooms, $\mathcal{G}$-JEPA without AD reaches $97.00\%$,
compared with $99.00\%$ for LeWM. Both methods achieve
near-saturated success on this reference set, with closely
matched success rates. We therefore focus the detailed
comparisons on the manipulation tasks.

% Section~\ref{sec:architecture_ablation} examines the contributions of grounding and action selection.

\begin{table}[t]
\centering
\small
\setlength{\tabcolsep}{3pt}
\caption{\textbf{Cross-method planning comparison.}  --- denotes not evaluated.}
\label{tab:cross_dataset}
\begin{tabular}{@{}lrrr@{}}
\toprule
Method & \multicolumn{1}{c}{PushT-Ada} & \multicolumn{1}{c}{OGBench Cube} & \multicolumn{1}{c}{TwoRooms} \\
\midrule
$\mathcal{G}$-JEPA w/o AD & $79.43_{\pm1.58}$ & $78.60_{\pm4.12}$ & $97.00_{\pm1.00}$ \\
$\mathcal{G}$-JEPA & $\mathbf{87.11}_{\pm1.17}$ & $\mathbf{100.00}_{\pm0.00}$ & --- \\
\midrule
LeWM & $56.90_{\pm2.35}$ & $72.80_{\pm2.53}$ & $\mathbf{99.00}_{\pm0.00}$ \\
TD-JEPA & $56.64_{\pm2.71}$ & $78.00_{\pm3.13}$ & --- \\
PSG-JEPA & $17.32_{\pm1.58}$ & $75.40_{\pm2.50}$ & --- \\
\bottomrule
\end{tabular}
\end{table}

\paragraph{Increasing goal distances.}
We evaluate long-horizon planning performance on tasks with increasing temporal separation between the initial and goal observations. Table~\ref{tab:goal_distance} reports results on PushT-Distance and PushT-Challenging. On PushT-Distance, $\mathcal{G}$-JEPA with and without AD exceeds LeWM at all three goal distances. Although adding AD slightly reduces success at 25 steps, it improves performance at both 50 and 75 steps. At 75 steps, the full planner reaches $55.21\%$, compared with $47.53\%$ without AD and $16.54\%$ for LeWM. PushT-Challenging shows the same trend, with the full planner reaching $52.22\%$ at 75 steps. These results show larger benefits from learned proposals at greater start--goal separations under this protocol.

\begin{table}[t]
\centering
\small
\setlength{\tabcolsep}{4pt}
\caption{\textbf{Planning with increasing goal distances.} We evaluate goal-reaching success with start--goal separations
of 25, 50, and 75 environment steps along recorded trajectories. }
\label{tab:goal_distance}
\begin{tabular}{@{}lrrr@{\hspace{12pt}}rrr@{}}
\toprule
& \multicolumn{3}{c}{PushT-Distance} & \multicolumn{3}{c}{PushT-Challenging} \\
\cmidrule(lr){2-4}\cmidrule(lr){5-7}
Method & 25 steps & 50 steps & 75 steps & 25 steps & 50 steps & 75 steps \\
\midrule
LeWM & $90.49_{\pm0.81}$ & $52.99_{\pm0.60}$ & $16.54_{\pm0.60}$ & $80.69_{\pm1.65}$ & $42.42_{\pm0.73}$ & $10.97_{\pm0.64}$ \\
$\mathcal{G}$-JEPA w/o AD & $\mathbf{96.09}_{\pm1.17}$ & $80.34_{\pm0.81}$ & $47.53_{\pm1.63}$ & $\mathbf{92.06}_{\pm2.38}$ & $75.92_{\pm1.00}$ & $44.03_{\pm1.73}$ \\
$\mathcal{G}$-JEPA & $91.80_{\pm1.17}$ & $\mathbf{85.55}_{\pm0.39}$ & $\mathbf{55.21}_{\pm0.90}$ & $83.33_{\pm2.38}$ & $\mathbf{82.30}_{\pm0.48}$ & $\mathbf{52.22}_{\pm0.96}$ \\
\bottomrule
\end{tabular}
\end{table}

\subsection{Ablation Study}
\label{sec:architecture_ablation}
\label{sec:planning_limits}

\paragraph{Grounding objectives.}
We ablate the training objectives to assess their contributions to planning performance. Table~\ref{tab:ablation_dataset} compares observed physical state grounding ($S$), nuisance invariance ($I$), and predicted physical state grounding ($P$), using Gaussian CEM throughout. The full training configuration ($S+I+P$) achieves $79.43\%$ success on PushT-Ada and $78.60\%$ on OGBench Cube, compared with $56.90\%$ and $72.80\%$ for the respective baselines. Removing predicted physical state grounding reduces success on OGBench Cube to $76.20\%$, while PushT-Ada retains $79.43\%$. Using observation grounding alone yields $55.73\%$ on PushT-Ada and $75.40\%$ on OGBench Cube. These results support the combined grounding objective, with its components contributing differently across tasks.

\begin{table}[t]
\centering
\small
\setlength{\tabcolsep}{4pt}
\caption{\textbf{Grounding-objective ablations.} $S$, $I$, and $P$ denote observed state grounding, nuisance invariance, and predicted physical state grounding.}
\label{tab:ablation_dataset}
\begin{tabular}{@{}cccrr@{}}
\toprule
State ($S$) & Invariance ($I$) & Predicted state ($P$) & PushT-Ada & OGBench Cube \\
\midrule
\ding{55} & \ding{55} & \ding{55} & $56.90_{\pm2.35}$ & $72.80_{\pm2.53}$ \\
\ding{51} & \ding{55} & \ding{55} & $55.73_{\pm1.58}$ & $75.40_{\pm1.65}$ \\
% \ding{55} & \ding{51} & \ding{55} & $19.40_{\pm1.26}$ & $74.20_{\pm1.75}$ \\
\ding{51} & \ding{51} & \ding{55} & $\mathbf{79.43}_{\pm0.90}$ & $76.20_{\pm2.57}$ \\
\ding{51} & \ding{51} & \ding{51} & $\mathbf{79.43}_{\pm1.58}$ & $\mathbf{78.60}_{\pm4.12}$ \\
\bottomrule
\end{tabular}

\end{table}

\paragraph{Temporal modules.}
We compare temporal architectures for the temporal grounding
module to assess their effect on downstream planning.
Table~\ref{tab:temporal_architecture} evaluates S4D, GRU,
Causal Transformer, and Attention-only under the same
training objective in Eq.~\ref{eq:total_objective} and
comparable parameter budgets.
Each module processes the observation and prediction sequences separately with shared parameters. Our S4D-based grounding module contains 55.4K parameters, including the physical state estimator, and serves as the reference budget for configuring the alternatives. Across three training replicates on PushT-Ada, S4D achieves the highest mean success, $78.12\%$, compared with $69.14\%$ for GRU and $61.46\%$ for Causal Transformer. Appendix~\ref{app:temporal_modules} details the training protocol and parameter accounting.
% We compare temporal architectures for the QLPG module to assess their effect on downstream planning. Table~\ref{tab:temporal_architecture} evaluates S4D, GRU, Causal Transformer, and Attention-only with the same QLPG objectives. Each module processes the observation and prediction sequences separately with shared parameters. Our two-layer S4D module contains 55.4K parameters, which serves as the reference budget for configuring the alternatives. Across three training replicates on PushT-Ada, S4D achieves the highest mean success, $78.12\%$, compared with $69.14\%$ for GRU and $61.46\%$ for Causal Transformer. Appendix~\ref{app:temporal_modules} details the training protocol and parameter accounting.

% Our two-layer S4D grounding module contains 55.4K parameters. We configure the alternative temporal modules with comparable parameter counts to compare planning performance under a common capacity budget.
% Table~\ref{tab:temporal_architecture} evaluates S4D, GRU, Causal Transformer, and Attention-only under the same $S+I+P$ objectives. All modules preserve the corresponding observation history when processing predicted latents. Results on PushT-Ada are averaged over three matched training replicates. Appendix~\ref{app:temporal_modules} details the comparison protocol and parameter accounting.

\begin{table}[htbp]
\centering\small
\setlength{\tabcolsep}{8pt}
\caption{\textbf{Temporal grounding architectures.} Mean success on PushT-Ada across three training replicates with a fixed planning seed.}
\label{tab:temporal_architecture}
\begin{tabular}{@{}lrr@{}}
\toprule
Architecture & Grounding params. (K) & Success (\%) \\
\midrule
S4D & 55.431 & $\mathbf{78.12}_{\pm3.34}$ \\
GRU & 54.279 & $69.14_{\pm14.89}$ \\
Causal Transformer & 55.607 & $61.46_{\pm10.61}$ \\
Attention-only & 55.555 & $53.39_{\pm34.02}$ \\
\bottomrule
\end{tabular}
\end{table}

\begin{figure}[!ht]
    \centering
    \includegraphics[width=\linewidth]{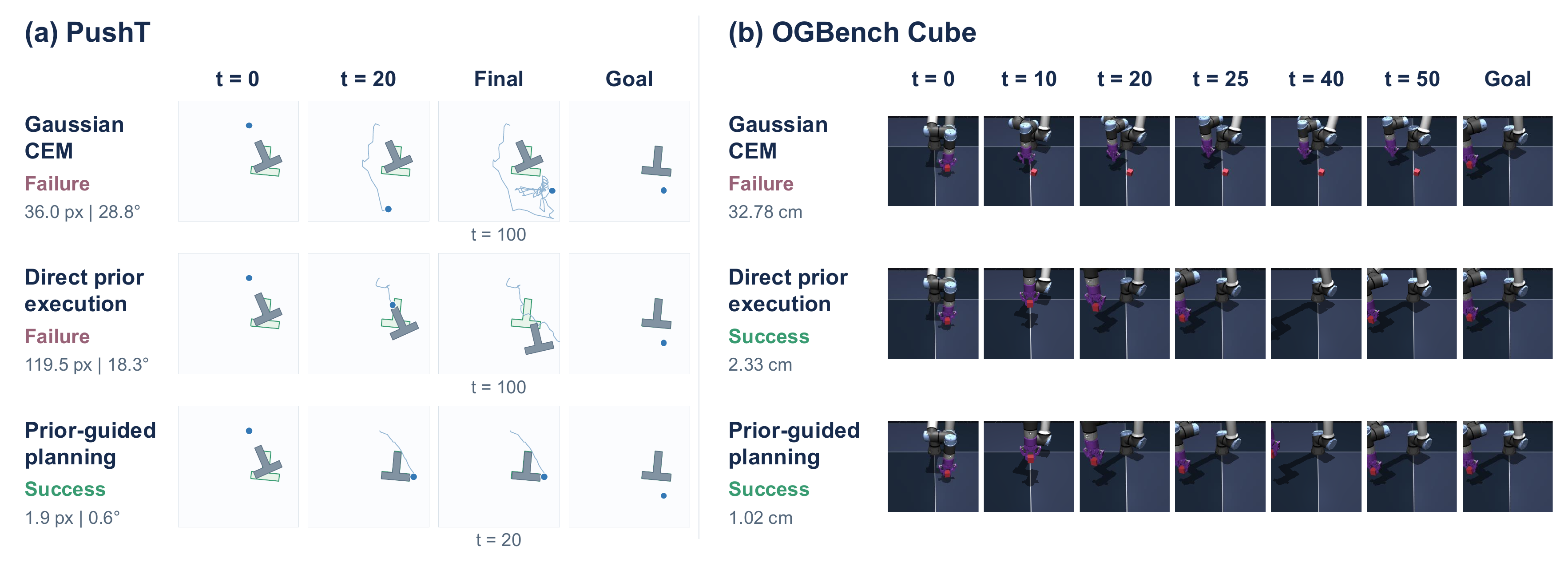}
    \caption{\textbf{Qualitative comparison of action selection.}
    Execution examples from the same initial state and goal
    within each task.
    (a) PushT shows recorded states, with goal poses in green
    and pusher paths in blue.
    (b) Cube shows simulator observations.
    Labels report success and final position/orientation errors
    (pixels/degrees) for PushT or cube--goal distance (cm) for Cube.
    $t$ counts environment steps, and \emph{Final} marks each
    method's stopping state in PushT.}
    \label{fig:planning_trajectories}
\end{figure}

\begin{figure}[t]
\centering
\begin{minipage}[t]{0.47\linewidth}
\vspace{0pt}
\centering
\includegraphics[width=\linewidth]{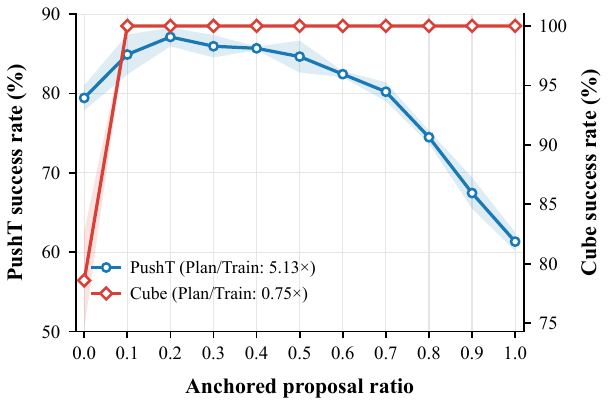}
\caption{\textbf{Anchored proposal ratio.} Mean success on PushT-Ada and OGBench Cube with $\pm$1 sample SD bands.}
\label{fig:diffusion_ratio}
\end{minipage}\hfill
\begin{minipage}[t]{0.52\linewidth}
\vspace{0pt}
\centering
\captionof{table}{\textbf{Action-selection ablations.} }
\label{tab:action_selection}
\small
\setlength{\tabcolsep}{3pt}
\begin{tabular}{@{}lrr@{}}
\toprule
Method & PushT-Ada & \shortstack{Cube} \\
\midrule
LeWM & $56.90_{\pm2.35}$ & $72.80_{\pm2.53}$ \\
\shortstack[l]{LeWM + AD w/o CEM} & $25.65_{\pm0.45}$ & $99.00_{\pm1.70}$ \\
\shortstack[l]{LeWM + AD} & $57.29_{\pm1.37}$ & $\mathbf{100.00}_{\pm0.00}$ \\
\midrule
$\mathcal{G}$-JEPA w/o AD & $79.43_{\pm1.58}$ & $78.60_{\pm4.12}$ \\
$\mathcal{G}$-JEPA w/o CEM & $25.91_{\pm1.37}$ & $99.80_{\pm0.63}$ \\
$\mathcal{G}$-JEPA & $\mathbf{87.11}_{\pm1.17}$ & $\mathbf{100.00}_{\pm0.00}$ \\
\bottomrule
\end{tabular}
\end{minipage}
\end{figure}

\paragraph{Action selection.}
We compare action-selection strategies to assess the benefit of combining learned proposals with model-based search. Table~\ref{tab:action_selection} compares Gaussian CEM, direct policy execution, and mixed-proposal CEM. The three $\mathcal{G}$-JEPA variants share the same frozen world model. The two variants using AD also share the same trained policy. Without CEM, one sampled action sequence is executed directly without world-model scoring. On PushT-Ada, mixed-proposal planning reaches $87.11\%$, compared with $79.43\%$ for Gaussian CEM and $25.91\%$ for direct execution. On OGBench Cube, direct execution already reaches $99.80\%$, leaving little room for the mixed planner's $100\%$. Direct policies trained on LeWM representations similarly reach $25.65\%$ on PushT-Ada and $99.00\%$ on OGBench Cube. The benefit of combining learned proposals with online search is more pronounced on PushT-Ada.
Figure~\ref{fig:planning_trajectories} illustrates this contrast: only prior-guided planning succeeds in the PushT example, while both direct prior execution and prior-guided planning succeed in Cube.

\paragraph{Proposal ratio.}
We vary the anchored proposal ratio to examine the balance between learned action proposals and Gaussian search. Figure~\ref{fig:diffusion_ratio} reports success rates with the world model, action prior, and total candidate budget fixed. On PushT-Ada, success peaks at $\rho=0.2$ with $87.11\%$ and decreases to $61.33\%$ at $\rho=1$, supporting a mixture of learned proposals and adaptive Gaussian search under this protocol. On OGBench Cube, every tested nonzero ratio reaches $100\%$, consistent with the high success of direct execution. To put this difference in context, the Plan/Train ratio divides median object displacement
in planning tasks by that in action-prior training windows.
It is $5.13$ for PushT-Ada and $0.75$ for OGBench Cube.
The larger displacement shift on PushT-Ada is consistent
with the benefit of retaining adaptive Gaussian search.
% the Plan/Train ratios in the legend compare median object displacement in planning tasks with that in AD training windows. On PushT-Ada, planning tasks require a median displacement $5.13$ times that of the training windows, whereas the ratio is $0.75$ on OGBench Cube. The larger displacement shift on PushT-Ada is consistent with the benefit of retaining adaptive Gaussian search, while the closer displacement scale on OGBench Cube is consistent with the strong performance of learned proposals.

\FloatBarrier

\section{Conclusion}
\label{sec:conclusion}
We introduced $\mathcal{G}$-JEPA, which combines temporal
physical grounding with prior-guided action planning.
A shared temporal physical grounding module aggregates temporal
context within observation and prediction sequences under
physical state supervision, while learned diffusions
proposals guide model-based online refinement through
mixed-proposal CEM.
% A shared QLPG module aggregates temporal context within
% observation and prediction sequences under physical state
% supervision, while learned diffusion proposals guide
% model-based online refinement through mixed-proposal CEM.
Experiments on PushT and OGBench Cube demonstrate improved
planning performance over the evaluated baselines.
Ablation studies support the effectiveness of temporal
grounding and the combination of learned proposals with
online search, while additional PushT evaluations demonstrate improved long-horizon planning performance.
Together, these results support integrating temporal physical
supervision and behavioral priors into JEPA-based planning.

\section*{AI Use Statement}
AI tools assisted with manuscript polishing and checking.
% ICLR 2027 requires an AI use statement; this section is excluded from the page limit.
% Official policy: https://iclr.cc/Conferences/2027/AIPolicyForAuthors

% \section*{Ethics Statement}
% Recommended when applicable; excluded from the page limit.

% \section*{Reproducibility Statement}
% Recommended by ICLR 2027; point to the appendix, anonymous code, configurations, seeds, and evaluation manifests.

% \subsubsection*{Author Contributions}
% Optional.

% \subsubsection*{Acknowledgments}
% Add only in the camera-ready version when needed for anonymity.

\bibliography{iclr2027_conference}
\bibliographystyle{iclr2027_conference}

\clearpage
\appendix

\section{Implementation and Evaluation Details}
\label{app:implementation}

\subsection{Data and Task Construction}
\label{app:data_tasks}

\textbf{PushT-Distance construction.}
A fixed random seed determines episode ordering and goal selection. Each selected episode supplies one shared goal and three starting contexts, ending 25, 50, and 75 environment steps before that goal. We select 256 distinct episodes with sufficient trajectory length and all three starting context windows in the validation split. Recorded actions are replayed to verify that each starting context reaches the shared goal within 10 pixels and $5^\circ$. Task selection is completed before model evaluation. PushT-Challenging filters out tasks that meet the success criteria at initialization, leaving 126, 209, and 240 tasks at the respective distances.

\textbf{Policy-data splits.}
PushT uses an episode-level 90/10 split for policy training
and validation. OGBench Cube uses 8,500 training and 500 validation episodes drawn from the 9,000 world-model training episodes. These policy splits are separate from the world-model splits specified in Section~\ref{sec:experimental_setup}.

\textbf{Physical targets and normalization.}
PushT targets comprise agent position and velocity, block position, and block angle, using LeWM's state and action normalization. OGBench Cube targets comprise six joint positions, end-effector position and yaw sine/cosine, gripper opening, contact, and cube position and yaw sine/cosine. OGBench Cube states and actions are standardized using statistics computed from the world-model training episodes. TwoRooms agent coordinates are divided by the image width 224.

\subsection{World-Model Training}
\label{app:world_training}

\textbf{Optimization.}
$\mathcal{G}$-JEPA uses AdamW with learning rate $5\times10^{-5}$, weight decay $10^{-3}$, and LeWM's warm-up/cosine schedule. PushT is trained for 20 epochs with a batch size of 32 and 10,000 mini-batches per epoch. OGBench Cube is trained for ten epochs with a batch size of 128. Loss weights are given in Section~\ref{sec:grounding_objectives}. 

\textbf{Grounding implementation.}
Each S4D layer produces 128-dimensional features for the
next layer or the physical state estimator. Training uses
four-frame windows, with encoded observations at times
0--2 and aligned one-step predictions at times 1--3.
The two branches share parameters and independently
aggregate temporal context from zero-initialized recurrent
states.

\textbf{Appearance perturbations.}
Across PushT, OGBench Cube, and TwoRooms, appearance augmentation is applied with probability 0.9 for nuisance-invariant representation learning. Brightness, contrast, and saturation factors are sampled from $[0.5,1.5]$ and shared across frames within each clip. Independent Gaussian pixel noise with standard deviation 0.05 is also added.
% For PushT, nuisance-invariant representation learning applies temporally shared brightness, contrast, and saturation factors from $[0.5,1.5]$, together with Gaussian pixel noise of standard deviation 0.05, with probability 0.9.

\textbf{Baseline adaptations.}
TD-JEPA's temporal-distance head is used during training but excluded from candidate scoring. For PushT, we add a task configuration to PSG-JEPA, which is not provided by its original implementation. All baseline results are obtained under our common planning protocols.

\subsection{Diffusion-Policy Training}
\label{app:diffusion_training}

\textbf{Architecture and optimization.}
For the policy defined in Section~\ref{sec:action_diffusion}, we use width 512, three FiLM residual MLP blocks, and a 64-dimensional diffusion-time embedding. The training target spans 25 environment actions. Both PushT and OGBench Cube policies use 100 forward diffusion steps, batch size 2,048, and 50 epochs. AdamW uses learning rate $10^{-4}$, weight decay $10^{-4}$, 1,000 warm-up updates, and cosine decay. Gradient clipping is set to 1 and exponential moving average decay to 0.995. Checkpoints are selected by validation noise-prediction loss. Both policies are trained using a fixed random seed. Policy-data splits are specified in Appendix~\ref{app:data_tasks}.

\subsection{Planning Protocols and Metrics}
\label{app:protocols}

\textbf{Cross-dataset planning settings.}
Within each dataset, planning comparisons use the same task sets, scoring rules, candidate budgets, and execution limits. Planning starts from one current image and retains up to three latents during recursive prediction. For PushT-Ada, the first five replanning rounds use terminal latent matching; later rounds weight goal-matching errors proportionally to $2^k$, $k=1,\ldots,K$. Table~\ref{tab:base_planning_protocols} summarizes the dataset-specific planning settings. Generating learned proposals incurs additional sampling cost beyond candidate evaluation.

\begin{table}[htbp]
\centering\small
\caption{Cross-dataset planning settings. CEM budgets denote candidates / elites / iterations.}
\label{tab:base_planning_protocols}
\label{tab:policy_protocols}
\begin{tabular}{lrrr}
\toprule
Setting & PushT-Ada & OGBench Cube & TwoRooms \\
\midrule
CEM budget & 100 / 10 / 10 & 300 / 30 / 10 & 300 / 30 / 30 \\
Executed environment steps / replan & 5 & 25 & 25 \\
Total environment-step budget & 100 & 50 & 50 \\
Position success threshold & 20 px & 4 cm & 16 px \\
Additional angle threshold & $20^\circ$ & --- & --- \\
\midrule
DDIM inference steps & 100 & 10 & --- \\
Anchored / Gaussian / mean candidates & 20 / 79 / 1 & 60 / 239 / 1 & --- \\
\bottomrule
\end{tabular}
\end{table}

\textbf{Success criteria.}
PushT requires both block-position and wrapped-angle errors to meet their thresholds at five-step execution endpoints. OGBench Cube and TwoRooms check object-position and agent-position errors, respectively, at environment steps. Tasks satisfying the success criteria at initialization are included in the evaluation, except in PushT-Challenging. PushT and TwoRooms are evaluated using three fixed planning seeds (42, 43, 44) shared across methods.

\textbf{Proposal-ratio sweep.}
We sweep $\rho$ from 0 to 1 in increments of 0.1 on the reported PushT-Ada and OGBench Cube tasks and planning seeds, holding the world model, action prior, candidate budget, and scoring rule fixed. We select $\rho=0.2$ after the sweep for $\mathcal{G}$-JEPA. It gives the highest mean on PushT-Ada; all nonzero ratios tie on OGBench Cube. Figure~\ref{fig:diffusion_ratio} shows the results.

\textbf{Direct-policy evaluation.}
Each frozen world model is paired with its own trained diffusion policy. Direct execution uses one sampled sequence per decision, without world-model scoring or CEM, on PushT-Ada and OGBench Cube, under the same execution limits and success criteria as the corresponding planners.

\textbf{PushT-Distance protocol.}
For the task sets constructed in Appendix~\ref{app:data_tasks}, planning uses $K=5,10,15$ for the 25-, 50-, and 75-step offsets, respectively, with 300 candidates, 30 elites, and 30 CEM iterations. All three distances use terminal latent cost, execute 15 environment actions per replan, and allow total budgets of 50, 100, and 150 actions. Success thresholds match PushT-Ada, with checks at initialization and five-step execution endpoints. All methods are evaluated using the fixed three planning seeds. The AD experiments use $\rho=0.2$, with 60 learned candidates, 239 Gaussian samples, and one retained mean candidate per iteration. For $K>5$, the action prior provides the first five action chunks of each learned proposal. The remaining chunks
are provided by CEM. Table~\ref{tab:goal_distance} reports results for both PushT-Distance and PushT-Challenging. PushT-Challenging results are computed from the same evaluated tasks after excluding those that satisfy the success criteria at initialization.

\subsection{Temporal-Module Comparison}
\label{app:temporal_modules}

\textbf{Training and evaluation.}
All modules in Table~\ref{tab:temporal_architecture}
are trained with the same objective in Eq.~\ref{eq:total_objective}. Each module processes observation and prediction sequences separately, with parameters shared between the two branches. S4D and GRU memories start from zero; attention operates causally within each sequence. Encoder initialization and data splits are matched within each replicate. Initialization/data seeds 3072--3074 are paired with training seeds 0--2. Training follows the PushT settings in Appendix~\ref{app:world_training}, with validation limited to 200 batches. Final checkpoints are evaluated on PushT-Ada using Gaussian CEM with a fixed planning seed (42), following the protocol in Appendix~\ref{app:protocols}.

\textbf{Parameter counts.}
The counts in Table~\ref{tab:temporal_architecture} include the temporal module and its physical state estimator, excluding the visual encoder and JEPA predictor. The alternative modules are configured to match the S4D grounding branch's parameter budget as closely as their architectures permit.

\end{document}

%% file: math_commands.tex
\usepackage{amsmath,amsfonts,bm}

\def\eqref#1{equation~\ref{#1}}
\def\1{\bm{1}}

\DeclareMathAlphabet{\mathsfit}{\encodingdefault}{\sfdefault}{m}{sl}
\SetMathAlphabet{\mathsfit}{bold}{\encodingdefault}{\sfdefault}{bx}{n}